\documentclass[sigconf]{acmart}

\copyrightyear{2021}
\acmYear{2021}
\setcopyright{acmcopyright}\acmConference[SIGIR '21]{Proceedings of the 44th
International ACM SIGIR Conference on Research and Development in Information
Retrieval}{July 11--15, 2021}{Virtual Event, Canada}
\acmBooktitle{Proceedings of the 44th International ACM SIGIR Conference on
Research and Development in Information Retrieval (SIGIR '21), July 11--15, 2021,
Virtual Event, Canada}
\acmPrice{15.00}
\acmDOI{10.1145/3404835.3462938}
\acmISBN{978-1-4503-8037-9/21/07}

\def\BibTeX{{\rm B\kern-.05em{\sc i\kern-.025em b}\kern-.08emT\kern-.1667em\lower.7ex\hbox{E}\kern-.125emX}}
    
\usepackage{svg}
\usepackage{amsmath}
\usepackage{import}

\usepackage{titlesec}
\titlespacing*{\subsection}{0pt}{0.3\baselineskip}{\baselineskip}

\usepackage{nicefrac}
\usepackage{siunitx}
\usepackage{array,framed}
\usepackage{booktabs}
\usepackage{
  color,
  float,
  epsfig,
  wrapfig,
  graphics,
  graphicx,
  subcaption
}
\usepackage{textcomp}
\usepackage{setspace}
\usepackage{latexsym,fancyhdr,url}
\usepackage{enumerate}
\usepackage{algorithm2e}
\usepackage{algpseudocode}
\usepackage{graphics}
\usepackage{xparse} 
\usepackage{xspace}
\usepackage{multirow}
\usepackage{csvsimple}
\usepackage{balance}
\usepackage{xcolor}

\usepackage{
  tikz,
  pgfplots,
  pgfplotstable
}
\usepackage{hyperref}

\usetikzlibrary{
  shapes.geometric,
  arrows,
  external,
  pgfplots.groupplots,
  matrix
}

\pgfplotsset{compat=1.9}

\usepackage{mathtools,}

\DeclareMathAlphabet{\mathcal}{OMS}{cmsy}{m}{n}

\DeclareGraphicsExtensions{%
    .png,.PNG,%
    .pdf,.PDF,%
    .jpg,.mps,.jpeg,.jbig2,.jb2,.JPG,.JPEG,.JBIG2,.JB2}

\begin{document}

\def\thetitle{DepressionNet: A Novel Summarization Boosted Deep Framework for Depression Detection on Social Media}
\title{\thetitle}






\author{$\text{Hamad Zogan}^{1,2}$, $\text{Imran Razzak}^{3}$, $\text{Shoaib Jameel}^{4}$, $\text{Guandong Xu*}^{1}$}
\affiliation{
  \institution{${}^{1}$School of Computer Science, University of Technology Sydney, Sydney \country{Australia}\\${}^{2}$College of Computer Science and Information Technology, Jazan \country{Saudi Arabia} \\ ${}^{3}$School of Information Technology, Deakin University, Geelong \country{Australia}, \\ ${}^{4}$School of Computer Science and Electronic Engineering, University of Essex \country{United Kingdom}}}
  \email{hamad.a.zogan@student.uts.edu.au, imran.razzak@deakin.edu.au, shoaib.jameel@essex.ac.uk, Guandong.Xu@uts.edu.au}
  




\renewcommand{\shortauthors}{Zogan, et al.}

\begin{abstract}
Twitter is currently a popular online social media platform which allows users to share their user-generated content. This publicly-generated user data is also crucial to healthcare technologies because the discovered patterns would hugely benefit them in several ways. One of the applications is in automatically discovering mental health problems, e.g., depression. Previous studies to automatically detect a depressed user on online social media have largely relied upon the user behaviour and their linguistic patterns including user's social interactions. The downside is that these models are trained on several irrelevant content which might not be crucial towards detecting a depressed user. Besides, these content have a negative impact on the overall efficiency and effectiveness of the model. To overcome the shortcomings in the existing automatic depression detection methods, we propose a novel computational framework for automatic depression detection that initially selects relevant content through a hybrid extractive and abstractive summarization strategy on the sequence of all user tweets leading to a more fine-grained and relevant content. The content then goes to our novel deep learning framework comprising of a unified learning machinery comprising of Convolutional Neural Network (CNN) coupled with attention-enhanced Gated Recurrent Units (GRU) models leading to better empirical performance than existing strong baselines.\let\thefootnote\relax\footnotetext{*Corresponding author.}
\end{abstract}

\ccsdesc[500]{Information systems}
\ccsdesc[500]{Information systems~Information retrieval}
\ccsdesc[500]{Information systems~Retrieval tasks and goals}
\ccsdesc[500]{Information systems~Clustering and classification}



\keywords {depression detection, social network, deep learning, machine learning, text summarization}

\maketitle

\section{Introduction}
\label{sec:intro}
Major depressive disorder (MDD), also known as depression, is among the most prevalent psychiatric disorders globally\footnote{https://www.who.int/news-room/fact-sheets/detail/depression} which leads to a substantial economic burden to the government. The problem also has a considerable impact on the living activities of an individual. It is associated with functional impairment, thus proving to be costly to society. Early determination and treatment of depression can help to improve the impact of the disorder on individuals. Sometimes early detection of depression is even more crucial for effective policing and to security agencies because it could adversely impact innocent citizens, e.g., mass shootings \cite{misperceptions2016mass} whose cause has been usually been attributed to mental health problems. As depression is a disorder which requires self-reporting of symptoms frequently, social media posts such as tweets from Twitter provide a valuable resource for automatic depression detection.



\begin{figure}
  \centering
  \includegraphics[scale=0.45]{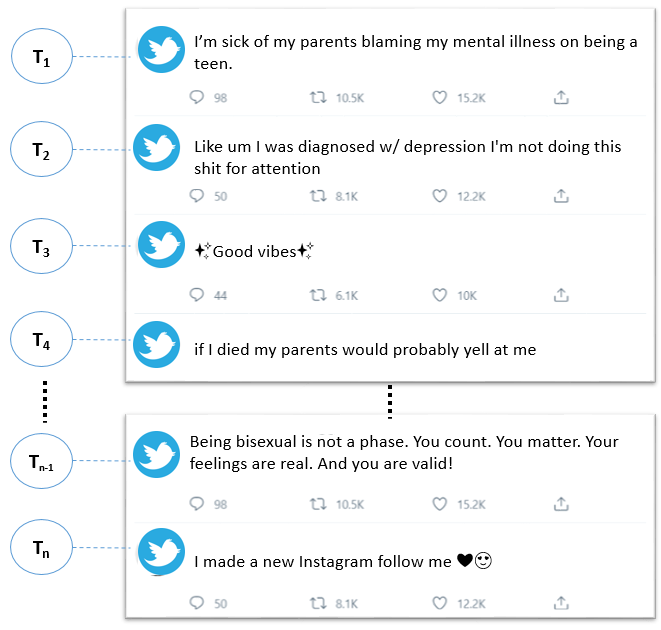}

      \vspace{-0.2cm}
           \caption{A sample of depressed user tweets. }
  \label{fig:dep_tweets}%
\end{figure}

People with depression often tend to hide their symptoms to avoid revealing that they are afflicted \cite{rodrigues2014impact} or they find it difficult to consult a qualified diagnostician \cite{KathleenSibelius}. Many go to online social media to express their underlying problems. One of the reasons is that they are willing to share their problems within their friend network thinking that they might offer help or advice. Sometimes they also implicitly leave clues which could point towards a state of depression or onset of early depression. Manually finding such users online by scanning through their posts will be very time-consuming. Therefore, the challenge is how we can propose effective computational techniques to automatically find such users online. Depression detection on Twitter is a promising and challenging research problem. Building an approach that can effectively analyze tweets with self-assessed depression-related features can allow individuals and medical specialists to understand the depression-levels of users. Researchers have proposed unsupervised methods to automatically detect depression online \cite{yazdavar2017semi, tsugawa2015recognizing} using computational techniques. One of the shortcomings in existing methods is that they tend to use every social media post of a user. We argue that this is not necessary because it tends to make the automatic depression detection system inefficient and even degrade the performance, e.g., dealing with ``curse-of-dimensionality'' and that these irrelevant posts may have a dominating impact more than depression-sensitive content. It is common for every user to share a varied set of posts online not just depression-related and we depict this user pattern through an example in Figure~\ref{fig:dep_tweets} which shows a variety of posts which might not be even relevant to depression. As a result, we need methods which could help dampen the impact of content which might eventually not help the classifier. We also need an effective feature selection strategy so that we could detect patterns from users which are implicit/latent which, unfortunately, cannot be modelled well by adopting simple term-frequency mining or modeling the surface-level features.



\begin{figure*}
  \centering
  \includegraphics[scale=0.3641]{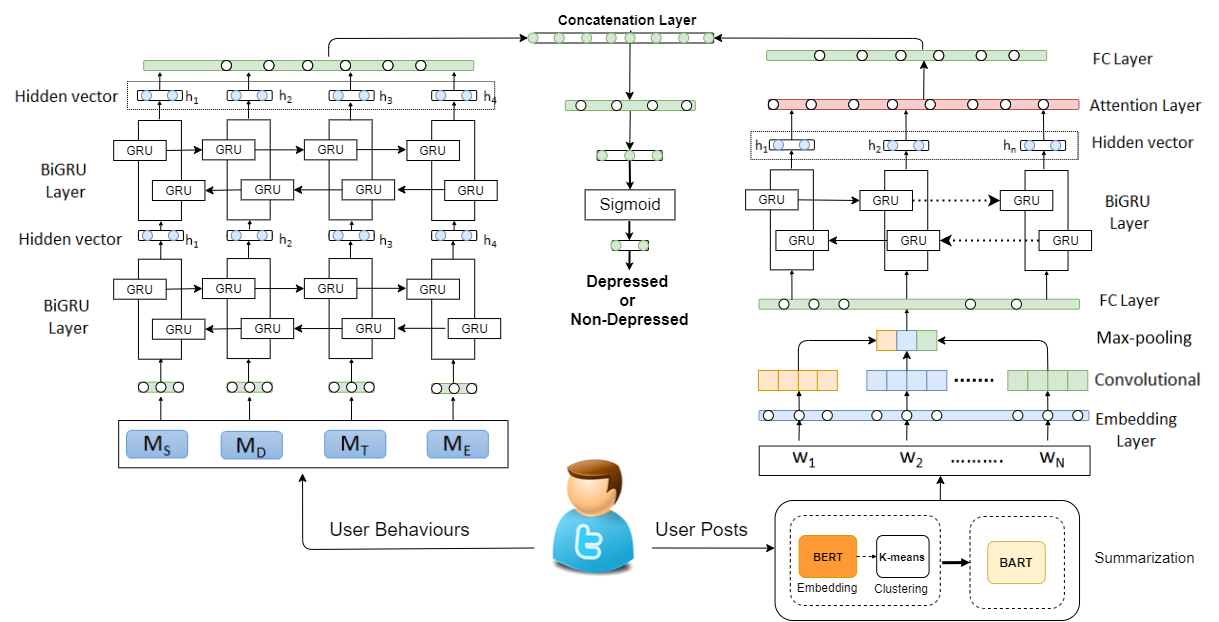}
  \caption{Proposed Framework (DepressionNet) for Depression Detection on Social Media }
  \label{fig:our_model}        
\end{figure*}

In single document automatic text summarisation \cite{tas2007survey}, the goal is to condense the document into a coherent content without losing any semantic information. Our proposed a novel computational framework trains on the content obtained after performing extractive-abstractive user-generated content summarization that helps select the most useful features for the classifier. The reason why we adopt the summarization approach is that it enables us to preserve the most salient content for every user and condenses them giving us a summary of the content. We use this summary in our novel deep learning framework which is based on Convolutional Neural Network (CNN) and Bidirectional Gated Recurrent Units (BiGRU) with attention. Exploiting the capability of the CNN network enables to model the features more faithfully. However, CNN usually performs suboptimally to capture the long term dependencies which are usually word order information. To mitigate this shortcoming, we introduce a bidirectional GRU model which belongs to the family of Recurrent Neural Network (RNN). We have used BiGRU to learn long-term bidirectional dependencies from backward and forward directions in our data because it has shown to perform well than a unidirectional GRU model. To further capture the user patterns, we have introduced various user behavioural features such as social network connections, emotions, depression domain-specific and user-specific latent topic information, and applied stacked BiGRU. A stacked architecture is obtained by having multiple BiGRUs runs for the same number of time steps which are connected in such a way that at each step the bottom BiGRU takes the external inputs. The higher BiGRU takes inputs externally as the input state output by the one below it. As pointed in \citet{he2016deep} residual connections between states at different depth are also used to improve information flow. Finally, we combine both user behaviour and post summarization in our framework, which we call DepressionNet, that consists of two shared hierarchical late fusion of user behaviour networks and a posting history-aware network. We conducted comprehensive experiments, and our model significantly outperforms existing approaches (+3\% in Acc and +6.5\% F-Score); thus, our \textbf{key contributions} are as follows:

\begin{enumerate}
  \item We propose a novel deep learning framework (DepressionNet) for automatic depression detection by combining user behaviour and user post history or user activity.
   \item We apply a abstractive-extractive automatic text summarization model based on the combination of BERT-BART that satisfies two major requirements:
   \begin{itemize}
       \item wide coverage of depression relevant tweets by condensing a large set of tweets into a short conclusive description.
       \item preserving the content which might be linked to depression.
   \end{itemize}
  \item To further make our prediction reliable, we have used information about user behaviour. To this end, we have developed a cascaded deep network that concatenates the behavioural features in different layers.
\end{enumerate}
\section{Related Work}
\label{sec:relwork}
Social media is one the platforms which could help discover and later propose ways to diagnose major depressive disorders. Researchers studied the effects of social media to predict depression since these platforms provide an opportunity to analyze individual user and state of mind and thoughts \cite{coppersmith2014quantifying,de2013predicting,de2014characterizing}. Digital records of people's social media behaviours, such as Facebook and Twitter, can measure and predict risks for different mental health problems, such as depression and anxiety. Additionally, these websites have been shown to allow machine learning and deep learning algorithms to be developed.



\subsection{Early Depression Detection}
\vspace{-0.3cm}
In the context of online social media, various studies in the literature have showcased that analysing user content and user textual information in social media has helped achieve some success for depression detection and mental illness. Lately, shared tasks such as CLEF eRisk \citet{losada2017erisk, losada2018overview} have included automatically detecting depression as early as possible from a users' posts. They discovered that there are features that capture differences between normal and depressed users. Several studies aim to analyze emotion, user network, user interactions, user language style and online user activities as features to identify depression on social media \cite{yazdavar2016analyzing,tsugawa2013estimating,shen2017depression,shen2018cross}. However, all these features are treated as an individual measurable property in different machine learning algorithms. For instance, to detect depressed users online \cite{de2013predicting} used the records of user activities with support vector machine (SVM) model for classifications and found the possibility to recognize depression symptoms through examining user activities on Twitter. In \citet{nadeem2016identifying} authors tried four different classifiers to classify user activity and found naive Bayes' model performed well. In \citet{tsugawa2015recognizing} authors also used user activities online; however, they determined that a detailed evaluation is needed to estimate the degree of depression through user activity history on social media.

One of the important mental health issues for new mothers is postpartum depression (PPD) \cite{miller2002postpartum,de2014characterizing,shatte2020social}. According to \citet{shatte2020social}, this depression could affect fathers too. The authors collected social media posts from fathers. They used the SVM model as a classifier and found that fathers could at risk of PPD. Other researchers study some other symptoms of depression, such as suicidal thoughts at an early stage. The authors in \cite{aladaug2018detecting} studied that people with suicidal thoughts leave a note on social media; therefore, they applied a classifier on posts, and the title on Reddit to differentiate suicidal and non-suicidal notes. They used term-frequency and inverse-document frequency (TF-IDF) as features and they found that logistic regression to be the ideal classifier for detecting the suicidal posts online.

Recently, some studies have started to target depressed user online, extracting features representing user behaviours and classifying these features into different groups, such as the number of posts, posting time distribution, and number followers and followee. In \citet{peng2019multi}, they extracted different features and classified them into three groups, user profile, user behaviour and user text and used multi-kernel SVM for classification. In \cite{shen2017depression} the authors proposed a multi-modalities depressive dictionary learning method to detect depressed users on Twitter. They extracted features from a depressed user and grouped these features into multiple modalities and proposed a new model based on dictionary learning that is capable of dealing with the sparse or multi-faceted user behaviour on social media.
 
The above-mentioned works have some limitations. They mainly focused on studying user behaviour than taking cues from user generated content such as the text they share which make it extremely difficult to achieve high performance in classification. These models also cannot work well to detect depressed user at user-level, and as a result, they are prone to incorrect prediction. Our novel approach combines user behaviour with user history posts. Besides, our strategy to select salient content using automatic summarization helps our model only focus on the most important information.

\subsection{Deep Learning for Depression Detection }\vspace{-0.3cm}
Use of a deep neural network (DNN) such as convolutional neural networks (CNNs) and long-short short-term memory (LSTMs) \citet{hochreiter1997long} have made notable progress in detecting mental illness on social media \cite{sun2019identification,umematsu2019improving,chen2018sentiment,rao2020mgl}. Deep learning has obtained impressive results in natural language processing (NLP) tasks such as text classification and sentiment analysis. Several works in the literature have concentrated on analysing user content and user textual information via deep learning models. For instance, \citet{shen2018cross} explore a challenging problem of detecting depression from two online social media platforms, Weibo and Twitter; therefore, they introduced the cross-domain DNN model using adaptive transformation \& combination features. Their model can address the heterogeneous spaces in various domains comprising of several features. To incorporate the users' behaviour, they extracted various features from both domains and classified them into four groups. Recently, different text classification models based on deep learning have been developed to determine if a single tweet has a depressive propensity.

In \cite{trotzek2018word} the authors targeted at the early detection of anorexia and depression on eRisk 2017 dataset, and in \cite{trotzek2018utilizing}, the authors find that CNN outperforms LSTM. They used the CNN model with GloVe \cite{pennington2014glove} and fastText \cite{bojanowski2017enriching} where for each record, they vectorized only the first hundred words. They have also trained other models using a broad range of features such as linguistic metadata and LIWS, and they found that linguistic metadata and word embeddings preformed well. This work inspired other researchers to classify posts on online forums to identify depression-related suicides \cite{yates2017depression}. For this task, they proposed a large-scale labelled dataset containing more than 116,000 users and proposed a model based on CNN architecture to identify depressed users concentrating on learning representations of user posts. Due to a large number of posts of each user, they created posts selection strategy to select which posts they used to train their model (earlier, latest and random posts), and among these three selection strategies, selection random posts gave better performance. On the other hand, RNN showed potential when it is applied to identify depression online. To classify depressed and healthy speech among patients screened for depression, \citet{al2018detecting} utilize an RNN model. They combined LSTM based model to concatenate the sequence and audio interviews of the patients. \citet{gui2019cooperative} fused two different data inputs, instead of text and audio. They observed that with the consideration of visual information of a post on Twitter, the meaning of that post is easy to determine. \citet{shen2017depression} proposed a model to detect depression using fusing images and the posts of a user in social media. To compute continuous representations of user sentences, they used Gated Recurrent Unit (GRU), and for images, they used 16-layer pre-trained Visual Geometry Group (VGGNet).


\subsection{Automatic Text Summarization}\vspace{-0.3cm}
Sequence-2-sequence models have been successfully applied to many tasks in NLP including text summarization \cite{nallapati2016abstractive,liu2018generating,tan2017abstractive}. Recently, transformer \cite{vaswani2017attention} model has become popular for automatically summarising text documents. \citet{liu2019hierarchical} proposed a model that can hierarchically encode multiple input documents and learn latent relationships across them. Recently, pre-trained language models have been commonly integrated into neural network models, such as BERT \cite{devlin2018bert} and BART \cite{lewis2019bart}. For tasks in NLP such as text summarization, these trained models have achieved state-of-the-art performance. In essence, the BERT model is based on an encoder-decoder network transformer. BERT has been adopted in the medical field as an extractive summarization approach, to summarise patients medical history \cite{vinod2020fine}. Recently the other pre-trained language model BART has attracted many researchers since it obtained new state-of-the-art performance in summarization task. BART contains a two-part bidirectional encoder and auto-regressive decoder. The robust of BART model lead \cite{gusev2020dataset} construct a corpus for text summarization in the Russian language, and they utilize a pre-trained BRT model for Russian language summarization. And their BART works extremely well, even though it was not originally intended for Russian language text summarization.

In this paper, we propose a novel depression detection model called (DepressionNet) using online user behaviours and summarization of his posting history. Our motivation for applying automatic text summarisation mainly comes from the fact that summarisation can aid our model to primarily focus on those content that is condensed and salient. To the best of our knowledge, we are the first ones to use automatic text summarization for depression detection on online social media.

\section{Our Novel DepressionNet Model}
\label{sec:methodology}
In this section, we present our proposed novel framework, DepressionNet for the task of automatic depression detection by fusing the user behaviour and the user post history. Figure \ref{fig:our_model} depicts the full model components. User posts can be abundant, redundant and may contain irrelevant information that might not give useful information to the computational model. This poses a significant challenge to effectively employ the knowledge learned from a user on social media which is already characterised by short and noisy content. Consider that a user $U_i$ has posts $[T_1,T_2,......,T_n]$ from the user activity history, where the total number of posts is $n$. Each $T_i$ is the \(i^{\text{th}}\) user-generated content. Our goal is to assign a label $y_i \in \{ \textit{depressed}, \textit{non depressed}\}$ to the user $U_i$ signifying a binary assignment whether the user is depressed or not depressed. To realise our goal, we have fused the user behaviour and user post history of each user $U_i$. The abstractive-extractive summarization can be defined as extracting and summarizing the user generated content $T_m$ from user history of $T_n$ tweets such that $T_m \leq T_n$. Besides, we incorporate user behavioural information social network, emotions, depression domain-specific and topic modelling denoted as $M_S, M_E, M_D$, and  $M_T$, respectively.
 



\subsection{Extractive-Abstractive Summarization}


\begin{figure}
  \centering
  \includegraphics[scale=0.45]{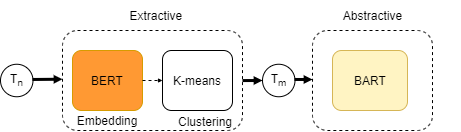}

      \vspace{0cm}
          \caption{Diagram to illustrate the process of user posts summarization.}
  \label{fig:summ_digram}%
\end{figure}
\begin{figure}%
    \centering
    \subfloat[\centering A depressed user posts before ]{{\includegraphics[width=4.3cm]{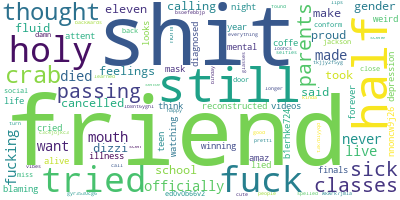} }}%
    \subfloat[\centering A depressed user posts after ]{{\includegraphics[width=4.3cm]{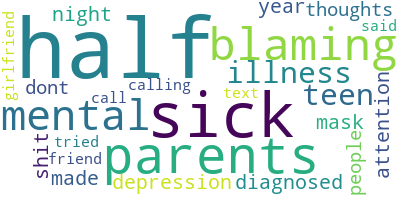} }}%
    
    \vspace{0.4cm}  
    \subfloat[\centering Non-depressed user posts before ]{{\includegraphics[width=4.3cm]{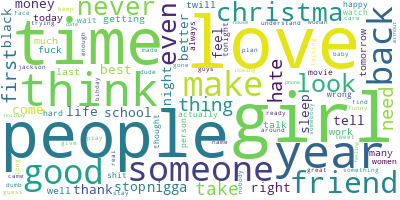} }}%
    \subfloat[\centering Non-depressed user posts after ]{{\includegraphics[width=4.3cm]{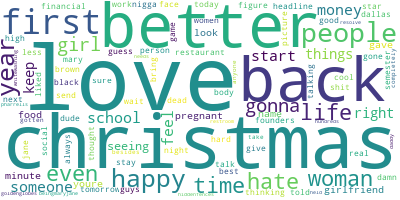} }}%
     \vspace{0.3cm}
    \caption{A word cloud depicting words from depressed and non-depressed users before and after extractive summarization. We show qualitatively that summarisation helps in selecting the most salient or focused content.}%
    \label{fig:cloud_dep}%
\end{figure}
\vspace{-0.3cm}
User post history plays a vital role in observing the progression of depression, thus, we considered the problem of depression detection by analyzing depressed user posts history to better understand the user behaviour for depression. Single-user post history summarization is the task of automatically generating a shorter representation of user historical post while retaining the semantic information. Figure~\ref{fig:cloud_dep} depicts the importance of text summarisation on a real user-generated content. We notice that a user who is depressed, the module distills the redundant and non-informative content, e.g., ``still'', ``eleven'' have been removed after summarisation from a depressed user and salient words such as ``sick'', ``mental'' have become prominent. We also observe the same pattern for non-depressed user where the focus only remains on most non-redundant patterns after summarisation. We argue that adopting summarisation technique is useful for the model to remove many irrelevant content so that we could focus on the most important information associated with a user.

We present a framework that incorporates an interplay between the abstractive and extractive summarisation. The reason why our model is based on extractive and abstractive framework is that extractive model helps automatically select the user generated content by removing redundant information. Abstractive framework further condenses the content while preserving the semantic content. Given that we might have a large amount of tweets associated with each user, this approach mainly helps reduce the amount of redundant information and noise in data. Our model relies on BERT-BART summarisation model which is a strong automatic text summarisation model based on contextual language models, which are very popular models for modeling text.


On the user posts ($T_n$), we have applied extractive tweets summarisation to select the most important tweets ($T_m$) from full set of user posts. The unsupervised extractive summary takes a pre-trained BERT model to perform sentence embedding \cite{miller2019leveraging}. Figure \ref{fig:summ_digram} depicts the design of our extractive-abstractive automatic summarization. As the BERT is trained as masked-language model, it gives us a vector representation ($[W_1,W_2....,W_j]_{T_i}$) for each tweet as the output which is grounded to tokens instead of sentences. We then perform \(k\)-means clustering on high dimensional vector representing semantic centers of text, the cluster centres are selected to extract most important tweets $T'_i$. For example, user \(A\) posts $n$ tweets $T=[T_1, T_2, T_3,....,T_n]$, the extractive summarization will return $m$ tweets $T'=[T_1, T_2, T_5 ....,T_{n-1}]$ whereas $T' \subseteq T$, i.e., in Figure \ref{fig:dep_tweets}, $T_3$ and $T_n$ are depression irrelevant tweets, extractive summarization excludes tweets $T_3$ and $T_n$.


Once, we have summarised the posts ($T'_m$) for each user ($U_i$) using the extractive summarisation, we can further condense the remaining redundant information that might gone undetected during the extractive summarisation phase. To this end, we apply the BART model which perform abstractive text summarisation. BART is denoising sequence-to-sequence autoencoder that uses transformer structure. The BART structure consists of two components: an encoder and a decoder. The encoder component is a bi-directional encoder compliant with the BERT architecture and the decoder component is an auto-regression decoder that follows GPT settings. We have used BART-large model \cite{lewis2019bart} which has originally been fine-tuned on CNN/DM dataset as abstractive summarization.




BART produces word embeddings that represents the summary of user posts at the word level. The word embeddings are then used as input to the stacked Convolutional Neural Network (CNN) and Bidirectional Gated Recurrent Units (BiGRU) with attention model to capture sequential information, such as the context of a sentence. The attention mechanism is advantageous in this scenario because the model helps focus on relevant words. The summary can be expressed as $S=\{w_{1},w_{2},..,w_{N}\}$, \noindent where \(S\) \(\in \,\,\mathbb {R}^{\mathbf {V}\times \mathit {N}}\), \(N\) represents the summary length, \(V\) represents the size of the vocabulary. Each word \(w_{i}\) in S is transformed to a vector of word \(x_{i}\) using Skipgram model available in word2vec library. We have used the pre-trained with 300-dimensional\footnote{https://github.com/mmihaltz/word2vec-GoogleNews-vectors} embeddings. The embedded summary sentence can be represented as:

\vspace{-0.2cm}
\begin{equation*}
\label{eq:emc2}
X=\{x_{1},x_{2},..,x_{N}\}  \tag{1}\end{equation*}

A weighted matrix of word vector will be utilized as embedded layer output and is input to the convolutional neural network (CNN) followed with layer of max-pooling and ReLU. The goal of CNN is to extract the most relevant embedded summary sentence features. The word vector representation of a tweet is typically complex, thus, the word vector  dimension is regularly taken by the CNN layer with kernel dimensions which extract important features by learning the spatial structure in summarized text through pooling. Finally, we add a fully connected (FC) layer which serves consolidated features to the BiGru.







\subsubsection{The Bidirectional Gated Recurrent Unit (BiGRU) Layer}
The resulting features from CNN layer are passed to the BiGRU, which is a RNN that can capture sequential information and the long-term dependency of sentences. Only two gate functions are used which are reset and update gates. Update gate has been used to monitor the degree with which the previous moment's status information has transported into the current state. The higher the update gate value, the more the previous moment's status information is carried forward. The reset gate has been used to monitor the degree with which the previous moment's status information is overlooked. The smaller the reset gate value, the more neglected the context will be. Both the preceding and the following words influence the current word in the sequential textual data, so we use the BiGRU model to extract the contextual features. The BiGRU consists of a forward GRU and a backward GRU that are used, respectively, to process forward and backward data. The hidden states obtained by the forward GRU and the backward GRU for the \( x_{t} \)  input at time \(t\) are forward(\( \ {h_{t}} \))  and  backward(\(  \ {h_{t}} \)), respectively.

\begin{align*} \text{forward}(\ {h_{1}})=& {GRU}(x_{t},\text{forward}(\ {h_{t-1}})) \tag{2}\\  \text{backward}(\ {h_{t}})=& {GRU} (x_{t},  \text{backward}(\ {h_{t-1}}))\tag{3}\end{align*}

The combination of the hidden state that is obtained from the forward GRU and the backward GRU \(\overrightarrow {h_{t}} \) and \( \overleftarrow {h_{t}}\) is represented as \( h_{t} \) as the hidden state output at time \(t\), and the output of the \(i^{th}\) word is \(h_{t}=\left((\text{forward}({h_{t-1}}) \oplus (\text{backward}({h_{t-1}})\right)\).

The attention mechanism helps the model to assign different weights to each part of the input and to reflect the correlation between features and performance results. Let \(H\) be a matrix consisting of output vectors [\(h_{1}, h_{2},h_{3}, . . . ,  h_{N} \)] which we obtain from BiGRU layer, where \(N\) here is the length of the summary sentence. The target attention weight \( u_{t} \) at timestamp \(t\) is calculated using the vectors \(h_{t} \) vector \(u_{t}=\tanh(h_{t})\).

We can get the attention distribution $a_{t}$, computed using a softmax function as \(a_{t}= \frac{\exp(u_{t})}{\sum\limits_{t=1}^{m}\exp(u_{t})}\). A user summarized posts attention vector $\bar{s}_{i}$ is calculated as the weighted sum of posts summarization features, using the dot product to sum products of $a_{t}$ and $h_{t}$ as follows: 
\(\bar{s}_{i}= \sum\nolimits_{t=1}^{m} a_{t}\bullet h_{t}\), where $\bar{s}_{i}$ is the learned features for the summary.

\subsection{User Behaviour Modelling}\vspace{-0.3cm}
\label{subsec:Behaviours}
\begin{table}
  \centering
  \small
      \caption{Summary of User Behaviour features}
      \scalebox{0.8}
      {
        \begin{tabular}{cp{17.175em}}
    Modality  & \multicolumn{1}{c}{Features Description} \\
    \midrule
    \multicolumn{1}{c}{\multirow{3}[2]{*}{Social\newline{}}} & 1-Posting time distribution for each user. \\
   & 2-Numeber of followers and friends (followee). \\
     {Network}  & 3-Number of tweets,  re-tweets and tweets length. \\
    \midrule
    \multirow{3}[2]{*}{Emotional} & 1-Number positive, negative or neutra emojies. \\
          & 2-(valance, arousal, dominance) score for each tweet  \\
          & 3-Calculate first person singular and plural from each tweet. \\
    \midrule
    \multicolumn{1}{c}{\multirow{2}[2]{*}{Domain\newline{} Specific}} & 1-Count of depression symptoms occurring in each tweet \\
          & 2-Count of antidepressant words occurring in each tweet \\
    \midrule
    Topic & 1-Using the LDA model \\
    \bottomrule
    \end{tabular}%
    }
    
  \label{tab:user_behav}%
\end{table}%

We have also considered user behaviour features and grouped all features into four types which are user social network, emotions, depression domain-specific and topic features obtained using a probabilistic topic model, Latent Dirichlet Allocation (LDA). We have shown more details in Table \ref{tab:user_behav}. These four feature types are also used in \cite{shen2017depression}. However, we have not considered \emph{User Profile feature} and \emph{Visual feature} types due to missing values. We have extracted these feature-types for each user that are described as below:

\noindent \textbf{I- Social Network Features}: We extracted several features related to user social interactions such as the number of followers and friends (followee). We have also considered user posting behaviour such as the number of tweets, retweets, and the length of tweets. Besides, we have extracted the posting time distributions as features, i.e., the number of tweets in an hour per day by user $U_i$. 

\noindent \textbf{II- Emotional Features}: User emotion plays important role in depression detection. We have considered emotional features such as valence, arousal, and dominance (VAD). The lexicon includes a list of English words and their valence, arousal, and dominance scores as features \cite{bradley1999affective}. We create a dictionary with each word as a key and a tuple of its (valance, arousal, dominance) score as value. We then parse each tweet and calculate VAD score for each tweet using this dictionary. We then add the VAD scores of tweets for a user, to calculate the VAD score. Tweets are rich in emojis and carry information about the emotional state of the user. Emojis can be classified as positive, negative or neutral, and can be specified as Unicode characters. For each of positive, neutral, negative type, first, we count their appearance in all tweets by user $U_i$.

\noindent \textbf{III- Domain-Specific Features}: The features which are often domain-dependent produce faithful classification results. We have considered two different types of domain-specific features which are depression symptoms and antidepressant related. For depression symptoms, we count the number of times any of the nine depression symptoms for DSM-IV criteria for a depression diagnosis \cite{edition2013diagnostic} is mentioned in tweets by user $U_i$. The symptoms are specified in nine lists, each containing various synonyms for the particular symptom. For each depression symptom, we count the number of times a symptom $S_j$ appeared tweets by user $U_i$. For antidepressant, we created a separate list of antidepressant medicine names from Wikipedia\footnote{https://github.com/hzogan/DepressionNet/blob/main/domain\_specific}, and count how many times the antidepressant name is mentioned by user $U_i$. 
 

\noindent \textbf{IV- Topic related Features}: Topic modelling uncovers the salient patterns (represented as distributions over the words in a tweet) by user $U_i$ under the mixed-membership assumptions, i.e., each tweet may exhibit multiple patterns. The frequently occurring topics plays a significant role for depression detection. We first consider the corpus of entire tweets of all depressed users and split each tweet into a list of words followed by assembling all words in decreasing order of their frequency of occurrence. We have removed stop words from the data. We applied unsupervised Latent Dirichlet Allocation (LDA) \cite{blei2003latent} to extract the latent topic distribution. For each user $U_i$, we compute, how many times each of words occurs in user's tweets separately.

To obtain fine-grained information, we have applied stacked BiGRU for the each of multi-modal features. In our experiments, we have considered two BiGRU that capture the behavioural semantics in both directions backwards and forward for each user followed by fully connected layer as shown in Figure \ref{fig:our_model}. Suppose the input which resembles a user behaviour be represented as $U_i$=[\(m_{1}, m_{2},m_{3}, . . . ,  m_{N} \) ] for \(i^{th}\) user. The outcome of behaviour modelling is the high-level representation that captures the behavioural semantic information and plays critical role in depression diagnosis (see ablation study section \ref{sec:results}). 

\subsection{Fusion of User Behaviour and Post History}\vspace{-0.3cm}
Figure \ref{fig:our_model} shows that the overall network consists of two asymmetric parallel networks (user post history network and user behaviour network) that consists of two shared hierarchical late fusion networks and a posting history-aware network that is combined with fully connected (FC) layer. The hierarchical temporal-aware network coalesces multiple fully connected layers to integrate user behavioural representation and user posting (history-aware  posting temporal network). For example, for user $U_i$, we have extracted a compact feature representing both behaviour and user posting history followed by a late fusion. The resulting framework models a high-level representation that captures the behavioural semantics. Similarly, the user post history comprises of representations extracted from user history that represent the gradual growth of depression symptoms. We have concatenated both representations to generate a feature map that considers both user behaviour and reflection of user historical tweets. The output of the DepressionNet network is a response map that denotes the similarity score between the depressed user and non-depressed user. As the network coalesces multiple hierarchical fully connected convolutional layers, thus, the network may have a different spatial resolution. To overcome this challenge, we exploit the max-pooling to down-sample the shallow convolutional layer to the same resolution as the deep convolutional layer. The hierarchical integration of user behaviour network results in a significant improvement of performance (see ablation study Section \ref{sec:results}).

\section{Experiments and Results}
\label{sec:eval}
In this section, we will describe our experimental setup in detail followed by comparisons with the state-of-the-art models.

\subsection{Baseline Methods}\vspace{-0.3cm}
We compare our proposed method with various strong comparative methods. Our comparative methods range from those that have been proposed for depression detection and for general text classification models because our setting also resembles that of binary text classification. For user behaviour features, we have used methods that have been applied for detection of mental illness. Multi-modal Dictionary Learning Model (MDL) has been proposed to detect depressed users on Twitter \cite{shen2017depression}. They used a dictionary learning to extract latent data features and sparse representation of a user. Support Vector Machines (SVM), is a popular and a strong classifier that has been applied on a wide range of classification tasks \cite{karmen2015screening} and it still remains a strong baseline. Na\"ive Bayes (NB) is a family of probabilistic algorithms based on applying Bayes' theorem with the ``naive'' assumption of conditional independence between instances \cite{deshpande2017depression,al2019depression}.
Given the popularity of contextual language models trained using modern deep learning methods, we have also investigated three popular pre-trained models, which are, BERT \cite{devlin2018bert}, RoBERTa \cite{liu2019roberta} and XLNet \cite{yang2019xlnet} for summarization sequence classification. We fine-tuned the models on our dataset. We also compare with GRU+VGG-Net with cooperative misoperation multi-agent (COMMA) \cite{gui2019cooperative}, where the authors proposed a model to detect depressed user through user posts (text) and images. They constructed a new dataset that contains users tweets with the images, based on the tweet ids \cite{shen2017depression}. 


\subsection{Dataset}\vspace{-0.3cm}
We have used a large-scale publicly available depression dataset proposed by Shen et al. \cite{shen2017depression}. The tweets were crawled and labelled by the authors. The dataset contains three components: \textbf{(1) Depressed dataset D1}, which comprises of 2558 samples labelled as depressed users and their tweets, \textbf{ (2) Non-depressed dataset  D2}, which comprises of 5304 labelled non-depressed users and their tweets. \textbf{ (3) Depression Candidate dataset D3}. The authors constructed a large-scale unlabeled depression-candidate dataset of 58810 samples. In our experiments, we used the labelled dataset: \textbf{D1} and \textbf{D2}. We preprocess the dataset by excluding users who have their posting history comprising of less than ten posts or users with followers more than 5000, or users who tweeted in other than English so that we have sufficient statistical information associated with every user. We have thus considered 4208 users (51.30\% depressed and 48.69 \% non-depressed users) as shown in Table \ref{tab:data}. For evaluation purpose, we split the dataset into training (80\%) and test (20\%) sets.

\begin{table}
  \centering
    \caption{Summary of labelled data used to train depression model} 
    \begin{tabular}{l|c|c}
    \toprule
    \toprule
    Description & Depressed & Non-Depressed \\
    \midrule
    Numer of users & 2159  & 2049 \\
    Number of tweets & 447856 & 1349447 \\
    \bottomrule
    \bottomrule
    \end{tabular}%
  \label{tab:data}%
\end{table}%

\subsection{Experimental Settings}\vspace{-0.3cm}
We have reported our experimental results after performing five-fold cross-validation. For summarization classification model (CNN-BiGRU with attention), the convolution layer, the window of size is set as 3, and the poling size for the max-pooling layer is set as 4. For BiGRU layer, we set the hidden layer to 32. For user behaviours representation model (stacked BiGRU), we used two layers of bidirectional GRU, and we set hidden neurons of each layer to 64.  All models are trained using the Adam optimizer \citet{kingma2014adam} using the default parameters: $\beta_1$ = 0.9,  $\beta_2$=0.999,  epsilon = 1e-7 and the learning rate =  0.001. We implement the extractive summarization through the clustering model based on Bert Extractive Summarizer \footnote{https://github.com/dmmiller612/bert-extractive-summarizer}. The number of topics in the topic model was set to 5 and from each topic 5 top words were chosen, which gave an overall better performance in our experiments after tuning. During the training phase, we set the batch size to 16. To evaluate the performance of the models, we have used accuracy, precision, recall, and F1 measure. In our framework, the depressed user classification was performed using dual-phased hybrid deep learning model (BiGRU + CNN-BiGRU with Attention)\footnote{https://github.com/hzogan/DepressionNet} for two attributes user behaviours and user posts summarization. To study the impact of different components in our model, we conduct ablation analysis.

\subsection{Results}\label{sec:results}\vspace{-0.3cm}
We evaluate the performance of user behaviour (Section \ref{subsec:Behaviours}) using stacked BiGRU model. We have used four different feature-types, social network, emotions, depression domain-specific and topic modelling, excluding user profile feature and visual feature due to missing values. Table \ref{tab:result_behaviours} shows the comparative results. We notice that behavioural features play an important role in the classification of depression. MLD achieved the second-best performance after Stacked BiGRU; however, stacked BiGRU performs the best to classify diverse features of user behaviours on social media. It outperforms the MLD model in precision by 4\%, recall, F1-score and accuracy, by 3\%, 3\% and 3\%, respectively.

\begin{table}
  \centering
        \caption{Effectiveness comparison different methods to detect depression via user behaviours.}
        \scalebox{0.7}
        {
    \begin{tabular}{cp{3.91em}p{3.91em}p{3.91em}p{3.91em}cccc}
    \toprule \toprule
    \multicolumn{1}{p{3.91em}}{\textbf{Model}} & \multicolumn{1}{p{3.91em}}{\textbf{Prec.}} & \multicolumn{1}{p{3.91em}}{\textbf{Rec.}} & \multicolumn{1}{p{3.91em}}{\textbf{F1}} & \multicolumn{1}{p{3.91em}}{\textbf{Acc.}} \\
    \midrule 
    SVM   & 0.724 & 0.632 & 0.602 & 0.644 \\
           NB    & 0.724 & 0.623 & 0.588 & 0.636 \\
           ~MDL  & 0.790 & 0.786 & 0.786 & 0.787 \\
           GRU   & 0.743 & 0.705 & 0.699 & 0.714 \\
           BiGRU & 0.787 & 0.788 & 0.760 & 0.750 \\
           Stacked BiGRU & \textbf{0.825} & \textbf{0.818} & \textbf{0.819} & \textbf{0.821} \\

    \bottomrule \bottomrule
    \end{tabular}%
}
  \label{tab:result_behaviours}%
\end{table}%

\begin{table}
  \centering
    \vspace{0.2cm}
    \caption{Comparison of different models for summarization sequence classification.} 
    \scalebox{0.7}
    {
    \begin{tabular}{p{9.545em}cccc}
    \toprule
    \toprule
    \textbf{Model} & \multicolumn{1}{p{3em}}{\textbf{Prec.}} & \multicolumn{1}{p{3em}}{\textbf{Rec.}} & \multicolumn{1}{p{3em}}{\textbf{F1}} & \multicolumn{1}{p{3em}}{\textbf{Acc.}} \\
    \midrule
    BiGRU (Att) & 0.861 & \textbf{0.843} & 0.835 & 0.837 \\
    CNN (Att) & 0.836 & 0.829 & 0.824 & 0.824 \\
    CNN-BiGRU (Att)  & 0.868 & 0.842 & 0.833 & 0.835 \\
    XLNet (base) & 0.889 & 0.808 & \textbf{0.847} & \textbf{0.847} \\
    BERT (base) & 0.903 & 0.770 & 0.831 & 0.837 \\
    RoBERTa (base) & \textbf{0.941} & 0.731 & 0.823 & 0.836 \\
    \bottomrule
    \bottomrule
    \end{tabular}%
    }
  \label{tab:summarization_seq}%
\end{table}%

\begin{table*}[!htb]
  \centering
  \caption{Comparison of depression detection performances in social media whence of four selected features.}
  \scalebox{0.7}
  {
    \begin{tabular}{lp{18.32em}cccc}
    \toprule
    \toprule
    \multicolumn{1}{p{8.635em}}{\textbf{Feature}} & \textbf{Model} & \multicolumn{1}{p{4.045em}}{\textbf{Precision}} & \multicolumn{1}{p{4.045em}}{\textbf{Recall}} & \multicolumn{1}{p{4.045em}}{\textbf{F1-score}} & \multicolumn{1}{p{4.045em}}{\textbf{Accuracy}} \\
    \midrule
     \multicolumn{1}{l}{\multirow{6}[2]{*}{User Behaviours}} & \multicolumn{1}{l}{SVM (\citet{pedregosa2011scikit})} & 0.724 & 0.632 & 0.602 & 0.644 \\
          & \multicolumn{1}{l}{NB (\citet{pedregosa2011scikit})} & 0.724 & 0.623 & 0.588 & 0.636 \\
          & \multicolumn{1}{l}{MDL (\citet{shen2017depression})} & 0.790 & 0.786 & 0.786 & 0.787 \\
          & \multicolumn{1}{l}{GRU (\citet{chung2014empirical})} & 0.743 & 0.705 & 0.699 & 0.714 \\
          & \multicolumn{1}{l}{BiGRU} & 0.787 & 0.788 & 0.760 & 0.750 \\
          & \multicolumn{1}{l}{Stacked BiGRU} & 0.825 & 0.818 & 0.819 & 0.821 \\
    \midrule
    \multicolumn{1}{p{8.635em}}{posts + Image} & GRU + VGG-Net + COMMA (\citet{gui2019cooperative}) & 0.900 & 0.901 & 0.900 & 0.900 \\
    \midrule
    \multicolumn{1}{l}{\multirow{6}[2]{*}{Posts Summarization}} & XLNet (base) (\citet{yang2019xlnet})  & 0.889 & 0.808 & 0.847 & 0.847 \\
          & BERT (base) (\citet{liu2019roberta}) & 0.903 & 0.770 & 0.831 & 0.837 \\
          & RoBERTa (base) (\citet{liu2019roberta}) & \textbf{0.941} & 0.731 & 0.823 & 0.836 \\
          & BiGRU (Att) & 0.861 & 0.843 & 0.835 & 0.837 \\
          & CNN (Att) & 0.836 & 0.829 & 0.824 & 0.824 \\
          & CNN-BiGRU (Att)  & 0.868 & 0.843 & 0.848 & 0.835 \\
    \midrule
    \multicolumn{1}{l}{\multirow{6}[2]{*}{Summarization + User Behaviures}} & CNN + BiGRU & 0.880 & 0.866 & 0.860 & 0.861 \\
          & BiGRU (Att) + BiGRU & 0.896 & 0.885 & 0.880 & 0.881 \\
          & CNN-BiGRU (Att) + BiGRU & 0.900 & 0.892 & 0.887 & 0.887 \\
          & BiGRU (Att) + Stacked BiGRU & 0.906 & 0.901 & 0.898 & 0.898 \\
          & CNN (Att) + Stacked BiGRU & 0.874 & 0.870 & 0.867 & 0.867 \\
          & \textbf{DepressionNet (Our Model)} & 0.909 & \textbf{0.904} & \textbf{0.912} & \textbf{0.901} \\
    \bottomrule
    \bottomrule
    \end{tabular}%
}
  \label{tab:dep_Comparison}%
\end{table*}%
For extractive summarization, we have used BERT and performed \(k\)-means clustering to select important tweets. We then applied the distilled version of BART (DistilBART) to extract abstract representation from these selected tweets. We used Distilbart provided by Hugging Face\footnote{\url{https://huggingface.co/transformers/model_doc/distilbert.html}}. To conduct abstractive summarization, we used different architectures such as XLNet, BERT, RoBERTa, BiGRU with attention and CNN-BiGRU with attention. From Table \ref{tab:summarization_seq}, we see that BiGRU-Att outperforms other models w.r.t recall, however, XLNet performs best among all the other models w.r.t accuracy and F1-\-score.




Tables \ref{tab:result_behaviours} and \ref{tab:summarization_seq} show that using only user posts summarization achieves a significantly better performance than using behaviours online. Consequently, using all attributes together further increases the performance of depression detection.





We have combined both summarization and user behaviour representation to capture various features using a hybrid deep learning model. Table \ref{tab:dep_Comparison} shows that our DepressionNet (Stacked\_BiGRU + CNN-BiGRU\_Att) model that show the best results compared with other models. Therefore, our proposed model can effectively leverage online user behaviour and their posts summarization attributes for depression detection. We further conducted a series of experiments and shown the performance of our model considering both qualitative and quantitative analyses. We have also compared the performance of our proposed framework with the state-of-the-art methods such as those based on user behaviour as shown in Table \ref{tab:result_behaviours}, post summarization as shown in Table \ref{tab:summarization_seq} and user selection of posts with images. Table \ref{tab:dep_Comparison} describes the comparative quantitative results. We can observe that the proposed framework DepressionNet performs better in comparison with comparative methods.
\begin{figure*}%
\small
    \centering
    \subfloat[\centering - Effectiveness of summarization sequence with features ]{{\includegraphics[scale=0.35]{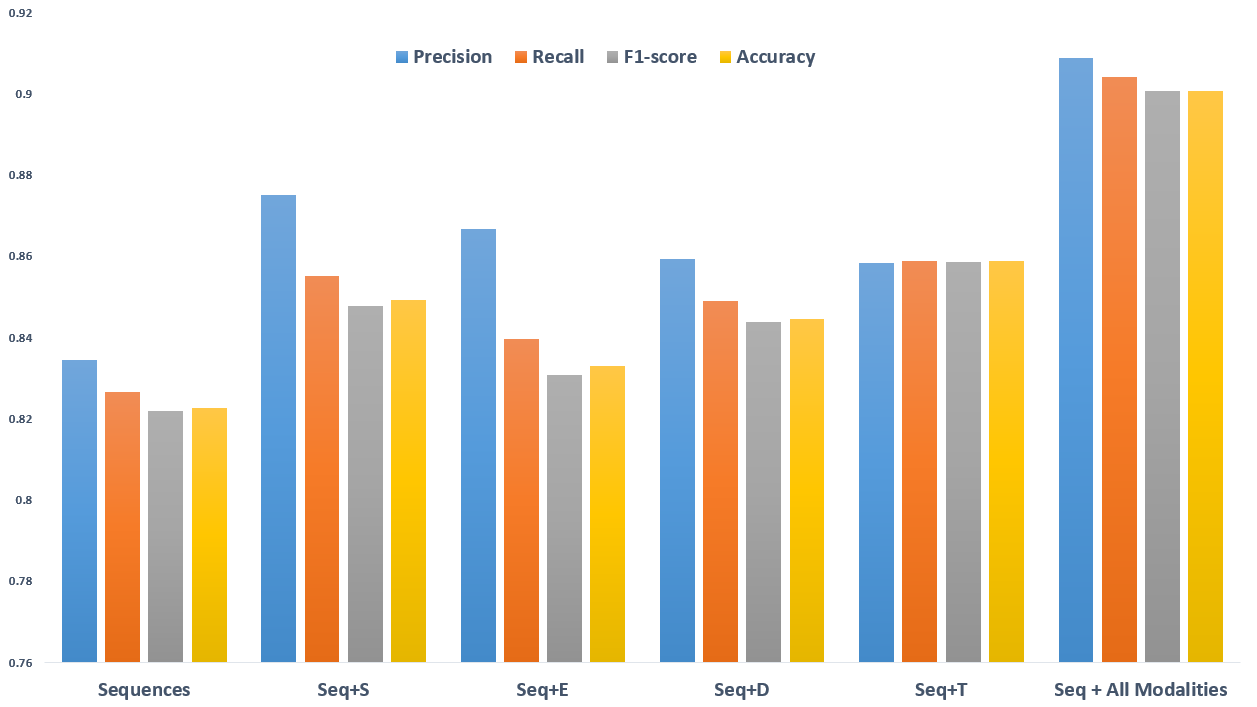} }}%
    \subfloat[\centering - Effectiveness of our model with features ]{{\includegraphics[scale=0.35]{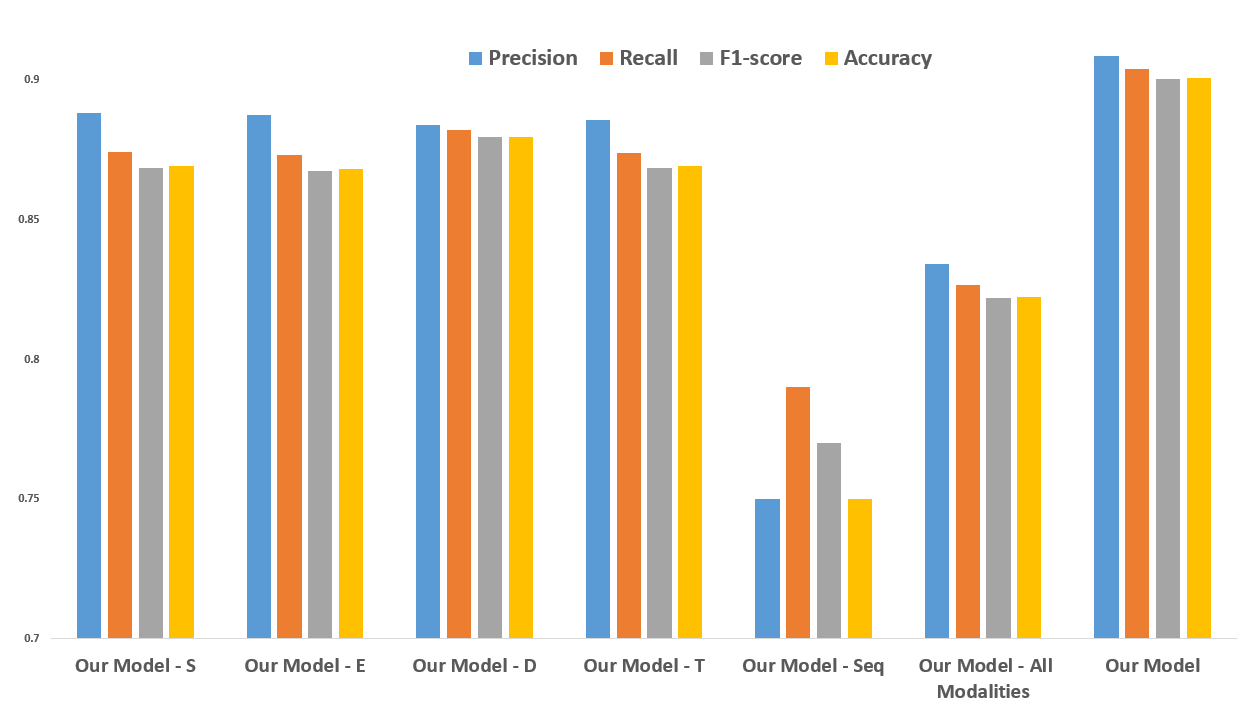} }}%
     \vspace{0.3cm}
            \subfloat[\centering - Model vs Text length.  ]{{\includegraphics[scale=0.40]{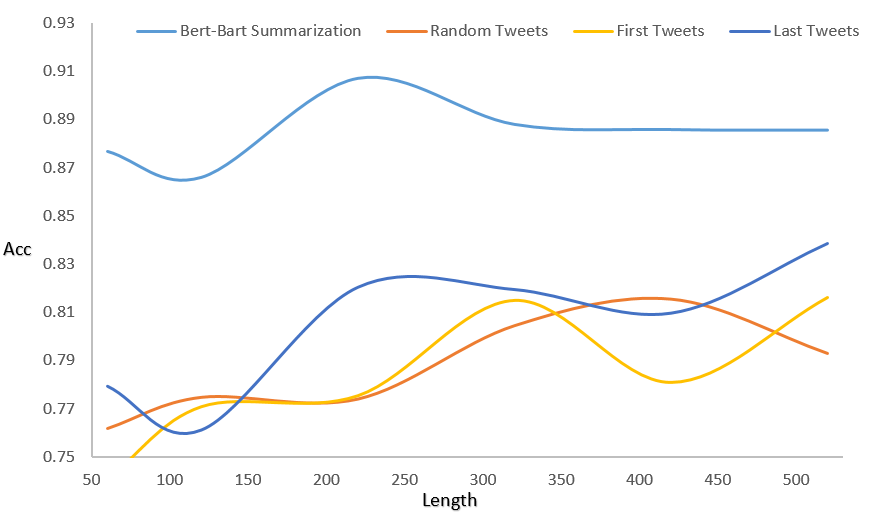} }}
         \subfloat[\centering -BART  ]{{\includegraphics[scale=.30]{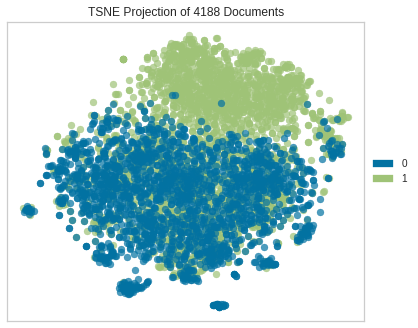} }}
    \subfloat[\centering - DistilBART  ]{{\includegraphics[scale=0.30]{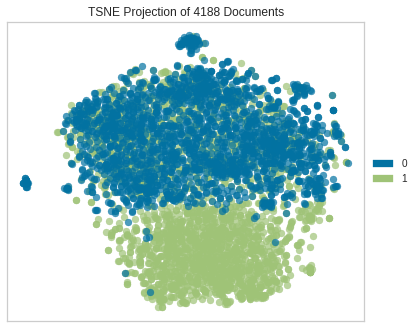} }}
\vspace{0.12cm}
   \caption{Experimental results: (a) Comparing the performance of our model by concatenating sequences with different feature types. (b) Showing the performance of our model by omitting different features. (c) our model performance vs Text length, with different inputs of data. (d) and (e) T-SNE visualization of Bart and DistilBart Summarization, where 1 represent depressed users posts and 0 represent non-depressed users posts}
   \label{fig: Model_results_performance}%
\end{figure*}

DepressionNet outperforms and improves by a percentage of ($\uparrow6.4\%$) on depression classification performance and achieves a score of 0.912 (F1-score) in comparison with comparative methods considering both user behaviour and post summarization which is shown in Table \ref{tab:dep_Comparison}. Gui et al., \cite{gui2019cooperative} showed that using images along with user posts plays an important role in depression detection. Similarly, the content of URLs may also be very helpful in the detection of depression as the patient often shares the link about the disease, and its medications among others. In the future, we plan to investigate the images as well as URLs along with post summarization and user behaviour. Our model also showed a slightly better performance in comparison to the model proposed in \citet{gui2019cooperative}. The improvement mainly comes from the interplay between the images the depressed user tweeted learnt jointly with text which helps give extra multi-modal knowledge to the computational model. 

Table \ref{tab:dep_Comparison} further shows comparative analysis. We can observe that not only DepressionNet but individual components of the proposed framework show a better performance in comparison to comparative methods. It is noticeable that our stacked BiGRU utilising the behavioural features, the social network, emotions, depression domain-specific and topic modelling outperformed the comparative methods based on leveraging just user behavioural patterns. Stacked BiGRU achieves a score of 0.819 (F1-score) in comparison to MDL and BiGRU. Table \ref{tab:result_behaviours} shows the detailed comparison of methods that comprise of the user behavioural patterns. We can derive similar conclusion considering accuracy, precision and recall. Similarly, our proposed extractive-abstractive summarization with BiGRU-attention showed better performance with 0.848 (F1-score) in comparison to 0.847, 0.831, 0.823, 0.835, 0.824 by XLNet, BERT, RoBERTa, BiGRU with attention and CNN with attention.

As depression is a gradually deteriorating disease, we considered the user historical tweets to better understand the user behaviour. As the depressed status intensifies, we can observe the evolving trajectory, however, such minor depression reflection could not be easily detected at early stage of depression, in comparison advanced stage. The selection of depression-related tweets and deep analysis of user history helps diagnose the patient at early stage. Based on experimental results, we have the following \textbf{observations}. 
\begin{enumerate}
    \item The consideration of user temporal tweet history helps detect the patient suffering from early-stage of depression.
    \item Extractive summarization selecting only important tweets results in not only improving the detection performance but also considerably reducing the computational complexity by filtering out unrelated tweets.
    \item  The hierarchical cascaded temporal-aware network coalesces multiple layers by integrating the user behavioural representation along with user tweets summary (history-aware temporal post summarization) which further can be used to analyze the progression of the disease and plan managements. 
   
    
    
\end{enumerate}

To further analyze the role played by each feature-type and contribution of the user behavioural attributes and user post summarization attribute, we removed the social network feature and denote this model as \emph{Our Model - S}, emotion feature and denote as \emph{Our Model - E}, domain-specific feature and denote as \emph{Our Model - D} and topic feature which we denote as \emph{Our Model - T}. We have also studied the performance of each attribute from our hybrid model separately. Specifically, we first removed the user behavioural multi-feature attribute (\emph{Our Model - All Feature}) and then removed user post summarization attribute (\emph{Our Model - seq}). 

We can see in Figure \ref{fig: Model_results_performance}(b) that our model performance deteriorates as we remove all the multi-features representing user behaviour attributes and degrades more without the post summarization. To understand the effectiveness of sequence attribute, we denote them respectively as following: \emph{Seq+S}, \emph{Seq+E}, \emph{Seq+D} and \emph{Seq+T}. As shown in Figure \ref{fig: Model_results_performance}(a), the combined sequence with topic feature contributes to depression detection slightly better than other features.


To further analyze the effectiveness of proposed summarization, we have also examined the DepressionNet model with four different data inputs such as first $m$ tweets ($U_i^F=[T_1,......,T_{20}]$), last $m$ tweets ($U_i^L=[T_{n-20},......,T_n]$) and random $m$ tweets ($U_i^R=Random_{20}[T_1,......,T_n]$). We can notice the significant superiority and stability of summarization approach over different input data as shown in figure \ref{fig: Model_results_performance}(c). In addition, we have also used t-SNE visualization to evaluate the abstractive summarization. We observe from Figure \ref{fig: Model_results_performance}(d) and (e) that we can see the distinct separation between the depressed summarization documents and non-depressed for both BART and DistilBART. 
\section{Conclusion}
\label{sec:conclusion}
We have proposed a novel hierarchical deep learning network that coalesces multiple fully connected layers to integrate user behavioural representation and user posting (history-aware posting temporal network). Our framework for depression detection works on online social media data which is characterised by first summarising relevant user post history to automatically select the salient user-generated content. Automatic summarization leads to a natural advantage to our computational framework which enables it to focuses only on the most relevant information during model training which significantly helps reduce the curse-of-dimensionality problem. Automatic summarization also introduces an additional benefit that there are no arbitrary design choices underlying our feature selection, e.g., discarding sentences with with certain predefined words or sentences within a certain length. As a result, our CNN-GRU model learns more discriminative attributes from data than the comparative models. The interplay between automatic summarization, user behavioural representation and model training helps us achieve significantly better results than existing state-of-the-art baselines. In the future, we will go beyond social media contents and use URLs, images as well as a mix of short and long user-generated content with traditional web pages. This would help give more contextual knowledge to the model that will help us focus on a task where our model not only detects depression but also automatically suggests possible diagnosis.
\section{Ethical Considerations}
This research intends to consider the needs of people who suffer from depression or depression-like symptoms on social media. 
  This study does not require  IRB/ethical approval. 

\label{sec:ethical}
\section{Acknowledgement}
This work is partially supported by the Australian Research Council (ARC) under Grant No. DP200101374 and LP170100891. Shoaib Jameel is supported by Global Challenges Research Fund (\#G004).

\bibliographystyle{ACM-Reference-Format}
\bibliography{main}


\end{document}